\documentclass{article}

\usepackage[preprint]{neurips_2026}

\usepackage[utf8]{inputenc}
\usepackage[T1]{fontenc}
\usepackage{hyperref}
\usepackage{url}
\usepackage{booktabs}
\usepackage{amsfonts}
\usepackage{amsmath}
\usepackage{amssymb}
\usepackage{mathtools}
\usepackage{nicefrac}
\usepackage{microtype}
\usepackage{xcolor}
\usepackage{graphicx}
\usepackage{subcaption}
\usepackage{multirow}
\usepackage{array}
\usepackage{enumitem}
\usepackage{algorithm}
\usepackage{algpseudocode}
\usepackage{xspace}
\usepackage{wrapfig}
\usepackage{needspace}

\newcommand{\method}{CLIFT\xspace}
\newcommand{\methodlong}{Conformal Lift-calibrated Introspective Filtering of Trajectories\xspace}
\newcommand{\ccc}{CCC\xspace}
\newcommand{\cegar}{CEGAR\xspace}
\newcommand{\aci}{ACI\xspace}
\newcommand{\vq}{VQ\xspace}
\newcommand{\vqs}{VQs\xspace}
\newcommand{\cts}{CTS\xspace}
\newcommand{\wai}{WAI\xspace}
\newcommand{\vwa}{VWA\xspace}
\newcommand{\om}{OM2W\xspace}

\title{CLIFT: Conformal Self-Verification for Web Agent Training and Test-Time Scaling}

\author{%
  Yifan Zhang \quad Yutong Dai \quad Viraj Prabhu \quad Zhiyuan Hu \quad Ran Xu \quad Zeyuan Chen\\
  Salesforce AI Research\\
  \small\texttt{\{\href{mailto:yifan.zhang@salesforce.com}{yifan.zhang}, \href{mailto:yutong.dai@salesforce.com}{yutong.dai}, \href{mailto:viraj.prabhu@salesforce.com}{viraj.prabhu}\}@salesforce.com}\\
  \small\texttt{\{\href{mailto:zhiyuan.hu@salesforce.com}{zhiyuan.hu}, \href{mailto:ran.xu@salesforce.com}{ran.xu}, \href{mailto:zeyuan.chen@salesforce.com}{zeyuan.chen}\}@salesforce.com}
}

\hypersetup{
  pdftitle={CLIFT: Conformal Self-Verification for Web Agent Training and Test-Time Scaling},
  pdfauthor={Yifan Zhang, Yutong Dai, Viraj Prabhu, Zhiyuan Hu, Ran Xu, Zeyuan Chen},
  pdfsubject={Web agent training and test-time scaling},
  pdfkeywords={CLIFT, conformal self-verification, web agents, trajectory selection}
}

\begin{document}

\maketitle

\begin{abstract}
Open-source web agents are now strong enough to execute realistic
browser tasks, but training them with reinforcement learning still
depends on weak supervision: binary task success is too sparse for
credit assignment, while frontier-language-model judges are too
expensive to call at every step and cannot be assumed available at
deployment. We introduce \method, a training and test-time scaling
method built around conformal self-verification. During training,
the agent answers natural-language verification questions about its
own rollouts; a Compositional Conformal Certifier keeps only
question signals whose URL-conditional evidence agrees with a
training-time judge, assigns signed trust weights through
polarity-aware lift, and blends the resulting verifier score into
per-step rewards in a way that never subtracts from the judge
baseline. At test time, the same certified bank is frozen and reused
as structured evidence for Conformal Trajectory Selection (CTS): the
agent samples a greedy rollout and one or more diverse retries, the
self-verifier summarises each URL trace, and a conservative
majority-vote rule chooses whether to swap away from the current
incumbent without calling any external judge. This single mechanism supports three
settings. On WebArena Infinity, \method\ achieves state-of-the-art
performance among open-source web agents. On VisualWebArena, a
bank trained with the open model transfers to GPT-5.5 at test time
and reaches state-of-the-art performance under the canonical
harness. On Online Mind2Web, without training an agent on the
benchmark, translating the certified question bank improves a
live-web agent in zero-shot evaluation. Together these results
position conformal self-verification as a way to turn costly judge
feedback into a reusable training signal and a judge-free
test-time scaling signal.
\end{abstract}

\section{Introduction}
\label{sec:introduction}

Open-source web agents have reached the point where they can operate
realistic browser interfaces, but the verifier remains the bottleneck
for improving them with reinforcement learning. A binary success
signal is easy to deploy but too sparse for credit assignment: when
all rollouts in a GRPO group~\citep{shao2024grpo} receive the same
outcome, the advantage estimate collapses. A frontier language model
judge~\citep{bai2022rlaif,liu2023geval} provides richer feedback,
but it is costly during training and cannot be assumed available at
test time. This tension is especially visible in web-agent benchmarks:
\wai~\citep{zhou2026webarenainfinity} uses programmatic per-app verifiers,
\vwa~\citep{koh2024visualwebarena} uses outcome-based URL/DOM
checks, and live-web suites such as \om~\citep{xue2025illusion}
depend on noisy LLM judges. The policy is often not the only limiting
object; the way we score and select its rollouts can dominate the
observed success rate.

\begin{table}[t]
\centering
\caption{\textbf{Feedback mechanisms for web-agent RL and test-time
scaling.} Dense = step-level feedback beyond binary outcome; URL =
URL-state grounding; Cert. = certified trust; J-free = judge-free test
time; TTS = test-time scaling; Xfer = reusable bank transfer.}
\label{tab:intro-comparison}
\footnotesize
\setlength{\tabcolsep}{4pt}
\renewcommand{\arraystretch}{1.02}
\resizebox{\linewidth}{!}{%
\begin{tabular}{@{}lcccccc@{}}
\toprule
\textbf{Approach} & \textbf{Dense} & \textbf{URL} & \textbf{Cert.} & \textbf{J-free} & \textbf{TTS} & \textbf{Xfer} \\
\midrule
Outcome RLVR / GRPO (trajectory outcome)~\citep{shao2024grpo}
  & $\times$ & $\times$ & $\times$ & \checkmark & $\times$ & $\times$ \\
Stratified GRPO (structural strata)~\citep{zhu2025stratifiedgrpo}
  & \checkmark & $\times$ & $\times$ & \checkmark & $\times$ & $\times$ \\
Rich-feedback self-distillation (teacher feedback)~\citep{hubotter2026sdpo}
  & \checkmark & $\times$ & $\times$ & \checkmark & \checkmark & $\times$ \\
Process / verifier reward models (learned PRM)~\citep{lightman2023letsverify}
  & \checkmark & $\times$ & $\times$ & \checkmark & $\times$ & $\times$ \\
Post-hoc search / BoN (sample then select)~\citep{snell2024testtime}
  & $\times$ & $\times$ & $\times$ & \checkmark & \checkmark & $\times$ \\
Conformal routing / judging (uncertainty gate)~\citep{badshah2026scope}
  & $\times$ & $\times$ & \checkmark & $\times$ & $\times$ & $\times$ \\
\midrule
\textbf{\method\ (URL-certified bank for training and CTS)}
  & \checkmark & \checkmark & \checkmark & \checkmark & \checkmark & \checkmark \\
\bottomrule
\end{tabular}}
\end{table}

\method\ asks whether the judge's feedback can be turned into a
reusable verifier that the agent carries with itself, rather than a
separate process reward model~\citep{lightman2023letsverify,wang2024math}
or a post-hoc best-of-$N$ scorer~\citep{cobbe2021training,snell2024testtime}.
\method\ maintains a bank of natural-language verification questions
about web-agent behaviour and certifies which questions are reliable
at which URL states. The same bank is used in two places
(Figure~\ref{fig:pipeline}). During training, certified bank evidence
is blended with the comparative judge to produce a denser per-step
reward and URL-stratified advantages. At test time, the trained agent
runs a greedy trajectory plus diverse retries; the frozen certified
bank summarizes each URL trace; and a conservative trajectory selector
keeps one emitted trajectory per task. The common object is not a new
model but a calibrated verifier:
a set of questions, URL scopes, polarity signs, and conformal trust
weights that can be reused after training.

\begin{wrapfigure}[16]{r}{0.60\linewidth}
  \vspace{-8pt}
  \centering
  \includegraphics[width=\linewidth]{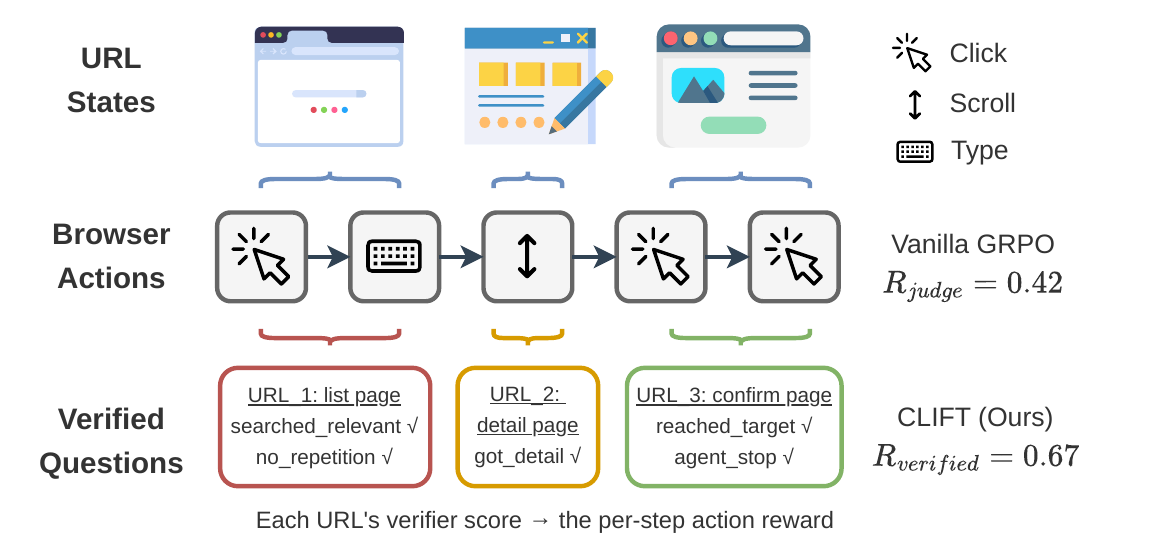}
  \caption{\textbf{\method\ concept.} Vanilla GRPO broadcasts one
  trajectory-scalar reward; \method\ scores each visited URL with
  verification questions and inherits the URL score to actions taken
  there. Full pipeline: Figure~\ref{fig:pipeline}.}
  \label{fig:concept}
  \vspace{-8pt}
\end{wrapfigure}

The certifier is the part that makes this more than prompt
engineering. Each verification question carries a URL scope and a
polarity sign indicating whether YES is evidence of progress or
failure. A Mondrian adaptive conformal tracker
\citep{gibbs2021aci,vovk2012mondrian} estimates whether that
question separates high- and low-judge rollouts at the relevant URL
granularity. Certified questions receive signed trust weights. During
training this gives a verifier reward that can only add evidence on
top of the judge baseline, preventing the reward-collapse failure of
earlier linear blends. During test-time scaling, the same signed
evidence is frozen and presented to a comparative selector over two
completed trajectories. In both cases, the question bank is a compact
interface between expensive judge feedback and the agent's own
rollouts.

We evaluate this idea in three complementary ways. On \wai, where
programmatic verifiers provide dense training supervision, \method\
improves a trainable open Gemma-4 agent and reaches the strongest
public 9-app score we are aware of; the point is both training and
CTS on the same open backbone. On \vwa, where the evaluation is more
outcome-based, we use the bank created during Gemma-4 training as a
test-time verifier for GPT-5.5, showing that the learned verification
knowledge transfers to a stronger closed backbone and reaches
state-of-the-art performance under the canonical harness. On \om, we
push transfer further: no agent is trained on \om; instead, the
certified question bank is translated and recalibrated, then used by
CTS to improve live-web test-time scaling. These three stories support
the same claim: conformal self-verification is useful both as a
training signal and as a reusable, judge-free test-time scaling signal.
Related work is discussed after the empirical section
(\S\ref{sec:relwork}).

\section{\texorpdfstring{\method: \methodlong}{CLIFT: Conformal Lift-calibrated Introspective Filtering of Trajectories}}
\label{sec:method}

\begin{figure}[t]
  \centering
  \includegraphics[width=\linewidth]{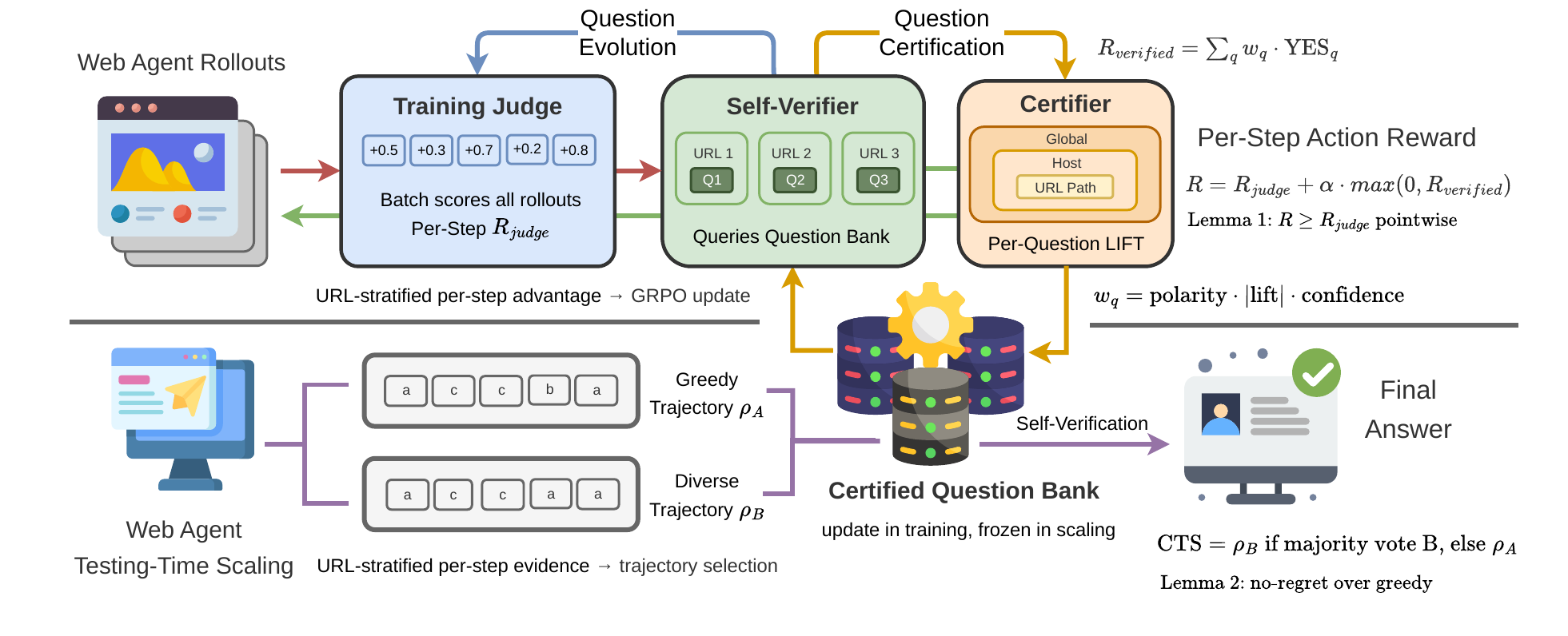}
  \caption{\textbf{\method\ overview.} \emph{Training:} rollouts
  are scored by a comparative judge and by a self-verifier that
  queries a polarity-tagged question bank. The \ccc\ certifies
  question weights by URL tier, yielding a signed verifier score
  that is asymmetrically added to the judge reward and then converted
  into URL-stratified token advantages. \emph{Test-time scaling:}
  the trained agent emits a greedy rollout and diverse retries; the
  frozen certified bank produces per-URL evidence for each pairwise
  comparison, and a conservative repeated-vote rule decides
  whether to keep the incumbent or swap.}
  \label{fig:pipeline}
\end{figure}

\method\ is a GRPO-family policy-gradient method
\citep{shao2024grpo} that changes how rollouts are scored and reused.
It turns natural-language verification questions into a conformally
certified URL-aware verifier, uses that verifier to add non-negative
per-step evidence during training, and freezes the same bank for CTS at
test time. Thus both halves of Figure~\ref{fig:pipeline} share one
object: questions with URL scopes, polarity signs, and certified trust
weights.

\subsection{The Conformal Verifier}
\label{sec:verifier}

A verification question (\vq) is a YES/NO check on the agent's
behaviour. Each question $q$ has a URL pattern $\pi_q$ and polarity
$\sigma_q\in\{-1,0,+1\}$ indicating progress, failure, or ambiguity.
At training and test time, the agent's base model answers applicable
questions with the task adapter detached. Answers at URL state $u$ are
aggregated into
\begin{equation}
\label{eq:signed-score}
R^{\vq}_i(u)
= \!\!\!\sum_{q \in Q^{\mathrm{cert}}(u)}\!\! w^q_{\tau^\star(u)} \cdot \mathbb{1}\!\big[y^q_i {=} \mathrm{YES}\big]
\;\in\; [-1, +1],
\end{equation}
where $Q^{\mathrm{cert}}(u)$ contains certified in-scope questions and
$w^q_{\tau^\star(u)}$ is the most-specific certified trust weight.
Positive weights indicate progress, negative weights indicate failure,
and an empty set gives $R^{\vq}_i(u)=0$.

\paragraph{Certification.}
The same question can be reliable on one app and useless on another, so
the \ccc\ keeps a Mondrian-\aci\ tracker
\citep{vovk2012mondrian,gibbs2021aci} for each question and URL tier
$\tau\in\{\text{global},\text{host},\text{URL-path}\}$. The
non-conformity score is polarity-aware:
\begin{equation}
\label{eq:nonconformity}
\begin{aligned}
s_t^{q,\tau} &= \mathbb{1}\!\left[\, \mathrm{pred}_q(y^q_t) \neq z_t\,\right],
\quad
z_t = \mathbb{1}[\hat R^{\mathrm{judge}}_t > \theta_j],\\
\mathrm{pred}_q(\mathrm{YES}) &= \mathbb{1}[\sigma_q {=} {+}1],
\quad
\mathrm{pred}_q(\mathrm{NO}) = \mathbb{1}[\sigma_q {=} {-}1].
\end{aligned}
\end{equation}
Positive questions conform when YES agrees with a high judge score;
negative questions conform when YES agrees with a low score. Neutral
questions do not certify. A tier $\tau$ of question $q$ is certified when
\begin{equation}
\label{eq:cert}
n^q_{t,\tau} \geq n_{\min}
\;\,\wedge\;\,
\alpha^q_{t,\tau} < \alpha^\star
\;\,\wedge\;\,
\big|\mathrm{lift}^q_{t,\tau}\big| > \mathrm{lift}_{\min}
\;\,\wedge\;\,
\sigma_q \neq 0,
\end{equation}
where $\mathrm{lift}^q_{t,\tau}$ is the mean judge-score difference
between YES and NO examples at tier $\tau$; $n_{\min}$,
$\alpha^\star$, and $\mathrm{lift}_{\min}$ are calibration
hyperparameters reported with the experimental setup. Certified
questions receive
\begin{equation}
\label{eq:trust-aggtrust}
w^q_\tau
= \sigma_q \cdot \big|\mathrm{lift}^q_{t,\tau}\big| \cdot \max\!\Big(0,\, 1 - \tfrac{\alpha^q_{t,\tau}}{\alpha^\star}\Big)
\;\in\; [-1, +1],
\end{equation}
with $w^q_\tau=0$ for uncertified tiers. At URL $u$ we query the most
specific certified tier, falling back URL-path $\to$ host $\to$ global.

\paragraph{Bank growth.}
The bank starts from a judge-grounded seed set of \vqs, each tagged with
a polarity. At periodic training intervals, a \cegar-style loop
\citep{clarke2000cegar} proposes a small batch of new URL-scoped
questions separating top and bottom rollouts. New questions enter
uncertified. A sanity veto demotes polarity to neutral when observed
lift contradicts the tag:
\begin{equation}
\label{eq:polarity-veto}
\sigma_q \leftarrow 0
\quad \text{if} \quad
n_q \geq n_{\mathrm{veto}}
\;\wedge\;
\mathrm{sgn}\!\big(\mathrm{lift}^q\big) \neq \sigma_q^{\mathrm{LLM}}
\;\wedge\;
\big|\mathrm{lift}^q\big| > \delta.
\end{equation}
The same lift estimate therefore certifies questions and catches
polarity mistakes.

\subsection{Training: Per-step Rewards and URL-stratified Advantages}
\label{sec:training}

For each task, one comparative judge call scores a GRPO group of $G$
rollouts and returns trajectory scores $R^{\mathrm{judge}}_i$ and
per-step scores $r^{i,\mathrm{judge}}_t$. In parallel, the
self-verifier answers bank questions for each rollout and URL state.
The training reward is
\begin{equation}
\label{eq:asym-blend}
\tilde r^i_t \;=\; r^{i,\mathrm{judge}}_t \;+\; \beta \cdot \max\!\big(0,\, R^{\vq}_i(u^i_t)\big),
\end{equation}
where $\beta$ controls the verifier contribution. Since the verifier
term is clipped at zero,
$\tilde r^i_t \geq r^{i,\mathrm{judge}}_t$: certified evidence can
strengthen the judge signal, but cannot drag it down.

\paragraph{URL-stratified per-token advantage.}
We discount step rewards into returns
$G^i_t = \sum_{\ell=t}^{T_i-1} \gamma^{\ell-t}\,\tilde r^i_\ell$
with discount $\gamma$, then group-normalise within each
\emph{(rollout-group, step-position, URL-state)} cell:
\begin{equation}
\label{eq:url-norm}
\widehat A^i_t = \frac{G^i_t - \bar G_{u^i_t}}{\sigma_{G, u^i_t} + \varepsilon}.
\end{equation}
Advantages are broadcast to tokens over tool-call spans; tokens outside
those spans receive zero gradient. The GRPO update is unchanged, but
credit assignment becomes URL-local.

\paragraph{Training-time judge.}
The judge sees compact traces with human-readable actions; opaque
browser-use object strings collapse judge variance. One prompt sees all
$G$ rollouts and returns trajectory scores, step scores, and
rationales for \cegar. The judge is training-only and is not called by
CTS.

\subsection{Test Time: CTS (Conformal Trajectory Selection)}
\label{sec:cts}

At deployment the external training judge is dropped. The policy emits
candidate rollouts under a small budget $K$: a greedy incumbent
$\rho_A$ and one or more diverse retries $\rho_B$. \method\ then uses
the frozen bank and base self-verifier in a pairwise selector to decide
whether to keep the incumbent or swap to the retry, avoiding brittle
step-wise selection.
For $K{>}2$, the same selector is applied as a tournament chain that
maintains one incumbent and still emits a single trajectory.

For each rollout $\rho \in \{\rho_A, \rho_B\}$, let
$U(\rho) = (u_0, u_1, \ldots, u_{T_\rho})$ be its sequence of distinct
URL states (consecutive duplicates collapsed). The verifier produces a
per-state evidence record at every $u_t \in U(\rho)$:
\begin{equation}
\label{eq:per-state}
E_t(\rho) = \Big\{\, \big(q,\; w^q_{\tau^\star(u_t)},\; \mathrm{lift}^q_{\tau^\star(u_t)},\; y^{q}(\rho, u_t)\big) \;:\; q \in Q^{\mathrm{cert}}(u_t),\; \pi_q \sim u_t \,\Big\},
\end{equation}
where $y^{q}(\rho,u_t)$ is the verifier's YES probability on the
partial trajectory. We keep this as structured evidence rather than a
single scalar.

The evidence sheets, action traces, and final states are passed to the
self-verifier in a comparative prompt, which outputs \textbf{A} (keep
greedy), \textbf{B} (swap), or \textbf{TIE}. This uses the bank as
structured evidence instead of a hand-designed URL-trace scalar.

To prevent verifier variance from causing harmful swaps, each pairwise
comparison uses a small odd number $m$ of independent self-verifier
calls and applies a conservative majority rule:
\begin{equation}
\label{eq:cts-decision}
\mathrm{CTS}(\rho_A, \rho_B) =
\begin{cases}
\rho_B & \text{if at least } \lceil(m+1)/2\rceil \text{ calls vote B}, \\
\rho_A & \text{otherwise (including ties and majority-A votes)}.
\end{cases}
\end{equation}
The asymmetric default is the safety valve: ties and uncertainty keep
$\rho_A$. If the bank-conditioned comparison is above chance, then
\begin{equation}
\label{eq:cts-noregret}
\mathbb{E}[\mathrm{SR}(\mathrm{CTS})] \;\geq\; \mathbb{E}[\mathrm{SR}(\rho_A)] - \epsilon(m, \delta),
\end{equation}
with $\epsilon$ decreasing as the majority vote becomes more reliable.
Here $m$ is the number of verifier votes in a pairwise comparison,
whereas $K$ denotes the rollout budget reported in the result tables.
CTS swaps only when evidence repeats; otherwise it defaults to greedy.

Training and test time therefore share one abstraction: per-URL-state
CCC-certified bank evidence. Training aggregates it into rewards and
advantages; CTS aggregates it through a comparative selector. The bank
is frozen after training but remains a reusable representation of what
the judge taught the agent to notice.

\section{Experimental Design}
\label{sec:experimental-design}

We evaluate \method\ as both a training method and a test-time
scaling method.

\begin{wraptable}[8]{r}{0.52\linewidth}
\vspace{-8pt}
\footnotesize
\centering
\caption{Experimental roles. \om\ supplies no reported policy-training data.}
\label{tab:exp-design}
\setlength{\tabcolsep}{3pt}
\renewcommand{\arraystretch}{1.04}
\begin{tabular*}{\linewidth}{@{\extracolsep{\fill}}lll@{}}
\toprule
Benchmark & Training role & Eval role \\
\midrule
\wai & train Gemma-4 & in-domain + board \\
\vwa & train bank & transfer to GPT-5.5 \\
\om & none & zero-shot transfer \\
\bottomrule
\end{tabular*}
\vspace{-8pt}
\end{wraptable}

\wai\ is the primary training setting: its self-hosted SaaS-style apps
expose programmatic verifiers that can support online RL and
public-board comparison. \vwa\ is the controlled transfer setting: its
Reddit, Shopping, and Classifieds environments use outcome-based
URL/DOM checks, so we test whether a bank learned while training an
open model remains useful when applied to GPT-5.5 at inference. \om\
is the live-web transfer setting: it contains 300 public-domain tasks
judged by an LLM rubric, and no reported policy is trained on \om\
data.

\subsection{Task synthesis and bank construction}
\label{sec:data}

Training tasks are synthesized without reading evaluation tasks. \wai\
seeds come from app descriptions; \vwa{} seeds come from SynthAgent~\citep{wang2026synthagent}
site exploration. Two frontier LLMs expand the seeds, and a shared
filter removes malformed, duplicate, hardcoded, pure-lookup, and
trivial-navigation tasks. The reported checkpoint trains on $1{,}886$
\wai{} tasks and $1{,}272$ \vwa{} tasks across $16$ LoRA shards. \om{}
is excluded from policy training and used only for bank transfer and
CTS. Appendix~\ref{sec:clift-training-setup} gives CLIFT training
details.

\subsection{Per-benchmark protocols}
\label{sec:benchmark-protocols}

The three benchmarks probe complementary properties of the same
verifier bank: \wai{} tests whether certified self-verification can
\emph{train} a stronger open agent, \vwa{} tests whether the
trained-on-open bank \emph{transfers} to a frozen closed backbone, and
\om{} tests whether the same bank \emph{recalibrates} for live-web
zero-shot selection.

\textbf{\wai.} We train Gemma~4~31B + LoRA with the full URL-path /
host / global CCC hierarchy. The comparison isolates three effects:
the base browser-use backbone, vanilla GRPO with the same comparative
judge but no certified bank, and \method{} with certified step-level
feedback. We then apply CTS to the trained checkpoint to measure
whether the frozen bank gives an additional no-regression test-time
selection gain on the same in-domain tasks.

\textbf{\vwa.} We use the bank learned during open-model training as
structured evidence for a stronger closed backbone. GPT-5.5 is not
updated and sees the same completed-task interface; the only added
component is CTS over finished trajectories. This setting isolates
whether the bank stores transferable web-interaction knowledge rather
than merely improving the Gemma-4 policy that produced it.

\textbf{\om.} We translate and recalibrate the \wai/\vwa{} bank for
live-web domains, then run CTS directly at inference. Reported rows
use upstream Gemma-4 or GPT-5.5 weights with no \om-specific LoRA.
Question translation rewrites benchmark-specific URL patterns into the
live-web equivalents, drops predicates that no longer apply, and adds
\om-specific failure-mode questions for bot-block pages, hallucinated
extraction, partial-filter application, and wrong-final-answer stops.
The recalibration pass scores each translated or new question on a
held-out \om{} rollout sample using the same lift-based certifier as
training, so any reported gain comes from cross-benchmark transfer of
the certified question taxonomy plus the calibration recipe, not from
policy training on \om{}.

\section{Result Analysis}
\label{sec:experiments}

We present results in the same order as the experimental design, from
the setting with the most direct supervision to the setting with no
policy training at all. \wai\ asks whether certified self-verification
improves an open trainable web agent; \vwa\ asks whether the bank
learned during open-model training remains useful for a stronger
closed model; \om\ asks whether the same bank can be translated and
recalibrated for live-web zero-shot selection.

\subsection{Training with certified feedback}
\label{sec:wai-results}

\begin{table}[t]
\centering
\caption{\textbf{\wai\ leaderboard slice.} Success rates on the public
9-app comparison set; values are percentages without percent signs.
The \method+\cts\ column reports the trained Gemma-4 backbone with
$K{=}3$ Conformal Trajectory Selection (one selected trajectory per
task). Overall is the task-weighted success rate over the 9 apps and matches the \method+\cts\
row of Table~\ref{tab:wai-app-ablation}.}
\label{tab:wai_public_baselines}
\small
\renewcommand{\arraystretch}{1.05}
\setlength{\tabcolsep}{3pt}
\begin{tabular}{lrrrrr}
\toprule
\textbf{} & \textbf{Gemini 3} & \textbf{Qwen 3.5+} & \textbf{Kimi} & \textbf{Gemma-4} & \textbf{Gemma-4 + \method+\cts} \\
\textbf{App} (n)              & \textbf{Flash + BU}  & \textbf{}       & \textbf{K2.5}      & \textbf{(base, ours)}    & \textbf{($K{=}3$, ours)} \\
\midrule
Elation Clinical (120)        &        82.0 &  54.0 &  50.0 &  69.2 & \textbf{91.7} \\
Elation Prescription (120)    &        81.0 &  42.0 &  23.0 &  70.0 & \textbf{91.7} \\
GitLab Plan \& Track (140)    &        64.0 &  37.0 &  39.0 &  59.3 & \textbf{69.3} \\
Gmail (60)                    &        75.0 &  57.0 &  70.0 &  60.8 & \textbf{78.3} \\
Handshake (200)               &        50.0 &  50.0 &  50.0 &  47.0 & \textbf{56.0} \\
Linear Acct Mgmt (120)        & \textbf{73.0} &  66.0 &  54.0 &  53.3 &        64.2   \\
PayPal Wallet (140)           &        89.0 &  71.0 &  71.0 &  82.1 & \textbf{90.7} \\
Super-human (120)             &        50.0 &  26.0 &  15.0 &  45.8 & \textbf{52.5} \\
Xero Invoicing (120)          & \textbf{81.0} &  56.0 &  52.0 &  75.0 &        79.2   \\
\midrule
\textbf{Overall (9 apps)} & 70.1 & 50.7 & 46.4 & 61.8 & \textbf{74.6} \\
\bottomrule
\end{tabular}
\end{table}

\begin{wrapfigure}[22]{r}{0.50\linewidth}
  \vspace{-6pt}
  \centering
  \includegraphics[width=\linewidth,height=0.145\textheight,keepaspectratio]{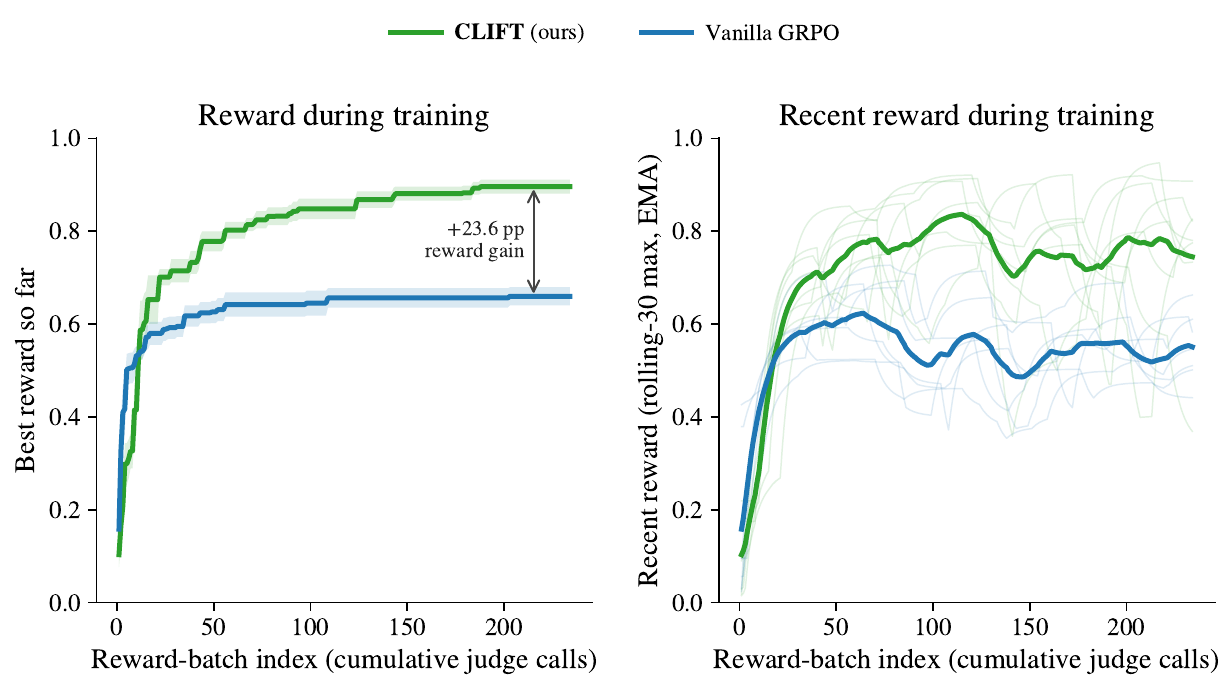}
  \caption{\textbf{\wai\ training reward.} \method\ separates from
  vanilla GRPO under the same judge and browser-use stack.}
  \label{fig:wai_reward_curves}
  \vspace{4pt}
  \includegraphics[width=\linewidth,height=0.105\textheight,keepaspectratio]{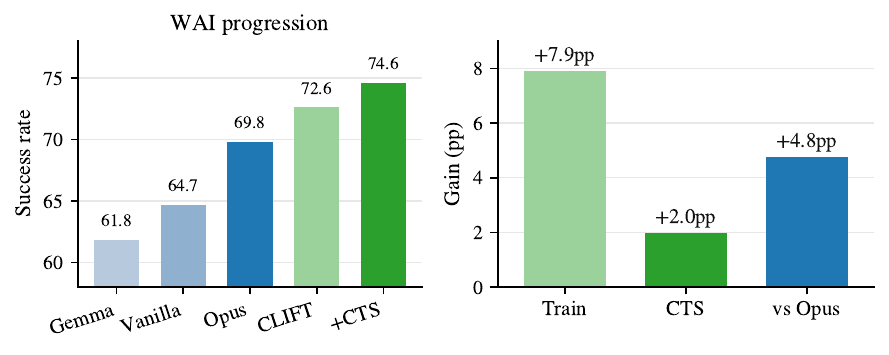}
  \caption{\textbf{\wai\ progression and gains.} Operating points and deltas.}
  \label{fig:wai_gain_decomposition}
  \vspace{-10pt}
\end{wrapfigure}

The public-board comparison in Table~\ref{tab:wai_public_baselines}
is the main \wai\ headline: \method-trained Gemma-4 with $K{=}3$ CTS
reaches the highest 9-app success rate among the published systems
we compare against (Claude Opus~4.6 is the training-time judge; CTS
itself calls no external judge). The gain is not concentrated in one
easy app: \method\ wins 7 of 9 apps, is competitive on the rest, and
improves the same Gemma-4 browser-use backbone by 12.8 percentage
points overall. The two apps led by Gemini~3~Flash are within the
leaderboard's integer-rounding noise, and the larger 110-task Gmail
variant is reported separately in Figure~\ref{fig:wai_gain_decomposition}.
This is the in-domain training claim: the verifier bank is not merely
a post-hoc reranker; it changes the online RL trajectory enough for a
31B open model to exceed the public frontier reference on the matched
\wai{} slice.

The training trace in Figure~\ref{fig:wai_reward_curves} explains why
the gap is not just a larger rollout budget at evaluation time. Once
the certified bank begins contributing signed per-step evidence,
training separates from vanilla GRPO before selection is ever used.
This lets the \wai{} story read from training dynamics to aggregate
success, rather than from a single final leaderboard number alone. The
reward curves also make the role of CTS easier to interpret: the
selector is operating on an already-improved policy, so a small
additional \wai\ lift is expected rather than a failure of test-time
scaling. For the aggregate comparison,
Figure~\ref{fig:wai_gain_decomposition} keeps the operating points in
order: base Gemma-4, vanilla GRPO, corrected Opus~4.6, \method, and
final \method+\cts. The same-denominator claim is no per-app
regression, so the CTS bar marks the final conservative selection
point rather than a pure reranking lift. Training supplies the large
\wai{} jump, CTS preserves it and adds a small safe lift, and the
larger test-time-scaling effects appear below.

\subsection{Bank transfer to a closed backbone}
\label{sec:vwa-results}

\begin{wraptable}[16]{l}{0.55\linewidth}
\vspace{-6pt}
\scriptsize
\centering
\caption{\textbf{\vwa{} success rates.} C/S/R are Classifieds,
Shopping, and Reddit; Avg is weighted by env size. For CTS rows, $K$
denotes rollout budget and one selected trajectory is emitted per task.}
\label{tab:vwa}
\setlength{\tabcolsep}{2.4pt}
\renewcommand{\arraystretch}{0.96}
\resizebox{\linewidth}{!}{%
\begin{tabular}{@{}lrrrr@{}}
\toprule
Method & C & S & R & Avg. \\
\midrule
SGV~\citep{andrade2026sgv} & $52.0$ & $\mathbf{57.0}$ & $33.0$ & $50.2$ \\
WALT~\citep{prabhu2025walt} & $\mathbf{64.1}$ & $53.4$ & $39.0$ & $52.9$ \\
\midrule
Gemma-4 base & $41.5$ & $28.5$ & $24.3$ & $30.9$ \\
Gemma-4 + \method{} & $44.9$ & $30.7$ & $31.0$ & $34.4$ \\
Gemma-4 + \method{}+\cts{} (K=2) & $45.3$ & $32.6$ & $31.4$ & $35.6$ \\
\midrule
GPT-5.5 + \method{} & $60.7$ & $43.1$ & $47.1$ & $48.5$ \\
\textbf{GPT-5.5 + \method{}+\cts{} (K=4)} & $62.4$ & $51.0$ & $\mathbf{50.0}$ & $\mathbf{53.7}$ \\
\midrule
Human & $91.7$ & $88.4$ & $87.1$ & $88.7$ \\
\bottomrule
\end{tabular}
}
\vspace{-4pt}
\end{wraptable}

\vwa{} is the controlled transfer benchmark between the two training
settings. Its three Docker environments have deterministic URL/DOM
outcome verifiers, but their visual grounding, page layouts, and
navigation routines differ from \wai{}'s SaaS apps. We therefore use
\vwa{} to ask whether a bank learned while training an open Gemma-4
agent can still provide useful evidence when the evaluated policy is
GPT-5.5, the executor weights are frozen, and only the certified
question bank is reused at inference. The published baselines in
Table~\ref{tab:vwa} follow the VisualWebArena comparison protocol
\citep{koh2024visualwebarena}; lower published baselines are omitted
from the main table for space.

Table~\ref{tab:vwa} follows the same accounting convention as the
\om{} table: each selected-output row emits exactly one trajectory per
task, so it is directly comparable to the base rollout on the same
task set. The Gemma rows isolate the training effect on an open
backbone: \method{} improves the base model before any closed-model
transfer is introduced, and CTS adds a smaller additional lift. This
mirrors the \wai{} pattern, but at lower absolute success rates because
\vwa{}'s visual and navigation demands are less aligned with the
training harness. The GPT-5.5 rows test the transfer claim directly.
The policy weights are unchanged; only the certified bank evidence and
CTS selection are added at inference. Adding CTS with rollout budget
$K{=}4$ raises the weighted average to $53.7$, narrowly above WALT's
$52.9$ reference, with gains on all three environments. The strongest
message is not that \vwa{} is solved (the human row remains far above
every agent) but that a bank constructed while training an open model
remains useful when the underlying executor is a closed model.

\subsection{Cross-benchmark transfer}
\label{sec:om2w-results}

\begin{wraptable}[14]{r}{0.52\linewidth}
\vspace{-4pt}
\footnotesize
\centering
\caption{\textbf{Transferred bank on \om.} Questions are reused from
\wai/\vwa{} or expanded, then recalibrated on \om{} rollouts; the final
certified bank has $75$ positive-lift and $25$ negative-lift questions
with mean lift $0.34$.}
\label{tab:om2w-transfer-summary}
\setlength{\tabcolsep}{2.6pt}
\renewcommand{\arraystretch}{1.02}
\begin{tabular*}{\linewidth}{@{\extracolsep{\fill}}llrr@{}}
\toprule
Origin & Construction & $|\mathcal Q|$ & Certified \\
\midrule
\wai{} bank & keep/rewrite & $36$ & $33$ ($91.7\%$) \\
\vwa{} bank & keep/rewrite & $48$ & $31$ ($64.6\%$) \\
\om{} expansion & live-web checks & $48$ & $36$ ($75.0\%$) \\
\midrule
\textbf{Final bank} & \textbf{132 qs} & \textbf{132} & \textbf{100 (75.8\%)} \\
\bottomrule
\end{tabular*}
\vspace{8pt}
\end{wraptable}

\om{} is our hardest setting: 300 live-web tasks across roughly 30
public domains, judged end-to-end by an LLM rubric rather than DOM
matching. We use it as a pure transfer benchmark: no reported policy
is fine-tuned on \om{} data. The only adaptation is at the verifier
layer: questions mined from the \wai/\vwa{} banks are filtered,
rewritten, expanded for live-web failure modes, and recalibrated on
\om{} rollouts. Table~\ref{tab:om2w-transfer-summary} compresses the
main bank evidence. The transferred taxonomy does not collapse under
the noisier live-web judge; instead, $100$ of $132$ questions certify,
including both positive progress signals and negative failure-mode
signals. This is the strongest evidence that the bank is a reusable
artifact rather than a benchmark-local prompt list.

\begin{wrapfigure}[10]{r}{0.45\linewidth}
\vspace{-12pt}
\centering
\includegraphics[width=\linewidth,height=0.12\textheight,keepaspectratio]{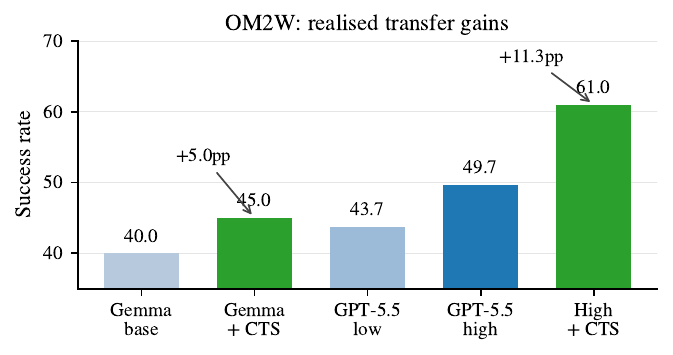}
\caption{\textbf{\om{} transfer gains.} Realised SR; $K$ is rollout budget.}
\label{fig:om2w-performance}
\vspace{-8pt}
\end{wrapfigure}

Figure~\ref{fig:om2w-performance} reports realised success: every
``+ ours'' row emits one selected trajectory per task, not an oracle
union over samples. On the open backbone, the transferred bank and
CTS raise Gemma-4 from $40.0$ to $45.0$, enough to exceed the GPT-5.5
low-effort baseline on the same judge ($43.7$). On the stronger
GPT-5.5 high-effort backbone, the same test-time mechanism raises
success from $49.7$ to $61.0$ at rollout budget $K{=}4$, a $+11.3$pp
gain. This is where test-time scaling is most visible: unlike \wai{},
where training already moves the greedy policy close to its ceiling,
\om{} leaves substantial room for the selector to choose among
distinct completed trajectories. The same numbers also show that the
open Gemma-4 backbone with transferred selection reaches $45.0$,
slightly above the GPT-5.5 low-effort baseline, even though no \om{}
policy weights are trained.

The cross-benchmark result also clarifies what transfers. The policy
does not: the Gemma rows use upstream weights and the GPT-5.5 rows
keep the closed model fixed. What transfers is the certified question
taxonomy plus the calibration recipe, which turns prior judge feedback
into structured evidence for a new live-web environment. Because the
bank is just a list of URL-scoped, polarity-tagged questions with
calibrated trust weights, recalibration on a new domain only requires
a sample of judge-scored rollouts; no policy update is needed to begin
selecting trajectories. Detailed bank construction and transfer tables
are in Appendix~\ref{sec:bank-transfer-appendix}.

\subsection{Ablation study}
\label{sec:ablations}

\begin{center}
\centering
\captionof{table}{\textbf{\wai{} ablation summary.} Success rates are shown
without percent signs. All training rows use Claude Opus~4.6 as the
comparative judge with the same browser-use backbone; only the bank/CTS
components vary. Steps are averaged over successful tasks in available
logs; CTS improves/ties/regresses 5/4/0 apps with 17.8\% average swap rate.}
\label{tab:wai-app-ablation}
\scriptsize
\setlength{\tabcolsep}{3.2pt}
\renewcommand{\arraystretch}{0.98}
\resizebox{\linewidth}{!}{%
\begin{tabular}{@{}llrrr@{}}
\toprule
Variant & Isolated change & SR & $\Delta$ vs base & Avg solve steps \\
\midrule
Gemma-4 base & no RL, greedy rollout & 61.8 & -- & 15.4 \\
Vanilla GRPO & scalar comparative-judge reward & 64.7 & +2.9 & 18.5 \\
\method{} & certified step reward + URL advantages & 72.6 & +10.8 & 16.0 \\
\method+\cts{} & add conservative trajectory selection & \textbf{74.6} & \textbf{+12.8} & 15.5 \\
\midrule
Opus~4.6 & closed-model reference, corrected common task IDs & 69.8 & +8.0 & -- \\
\bottomrule
\end{tabular}}
\end{center}

Table~\ref{tab:wai-app-ablation} consolidates the ablations. Base
Gemma-4 is the no-RL reference; vanilla GRPO isolates scalar
judge-reward training; \method{} adds certified step feedback; and
\method+\cts{} adds only test-time selection. Training accounts for
most of the WAI gain, while CTS improves five apps, ties four, and
never lowers the single-rollout \method{} anchor.

The decomposition further shows that the gain is not explained simply
by additional interaction or inference-time search. Vanilla GRPO
improves the base model by only 2.9 points while increasing the average
number of steps on successful trajectories from 15.4 to 18.5.
In contrast, \method{} reaches a 10.8-point improvement with an average
of 16.0 steps, suggesting that the certified URL-local feedback primarily
improves credit assignment rather than encouraging longer trajectories.
\cts{} then adds a further 2.0 points while reducing the average to 15.5
steps. Together with its 17.8\% swap rate and zero per-app regressions,
this supports the intended division of labor: \method{} produces the
stronger policy during training, while \cts{} intervenes selectively at
test time when the certified bank provides sufficient evidence to prefer
an alternative rollout.
\section{Discussion and future work}
\label{sec:discussion}

\method's failures are intended to be visible. On \wai\ and \vwa,
many URL states accumulate enough observations for signed trust
weights; on \om, the noisier live-web judge leaves more questions
uncertified. This refusal is a diagnostic rather than a silent collapse:
it says the current judge, state partition, or question bank does not
support reliable distillation, and keeps noisy verifier evidence out of
both the training reward and the CTS selector.

The study is still bounded. We train one open-weight backbone and rely
on a frontier judge during training and bank construction; if that judge
is systematically biased, calibration can detect disagreement only
through observed data. The transfer results show that the bank can help
stronger closed-source agents, but not that one bank is universal across
all web domains. Stronger open web agents also carry misuse risk, so
deployment should retain rate limits and domain-specific safeguards.

A natural next step is to replace natural-language questions with
executable browser-state assertions written by a coding agent. The
agent would learn not only to act on the web, but also to write compact
checks that can be executed during rollouts and reused for test-time
selection, reducing dependence on a separate self-verification bank
while preserving auditability.

\section{Related work}
\label{sec:relwork}

CLIFT sits at the intersection of web-agent benchmarks, RL credit
assignment, test-time search, and reliable verifier-based selection. The
closest systems usually improve one component of this loop: stronger web
interfaces for agents, denser reward signals for training, or calibrated
decision rules for deployment. CLIFT instead studies how one certified
verification bank can connect these components, carrying judge feedback
from online RL into test-time trajectory selection.

\paragraph{Web agents and test-time search.}
Language agents combine reasoning, browser actions, and tools
\citep{nakano2021webgpt,yao2023react,shinn2023reflexion,schick2023toolformer,qin2023toolllm};
search methods add deliberation or tree expansion at inference time
\citep{wei2022cot,yao2023tot,zhou2024lats,snell2024testtime,song2024trialanderror};
and modern web, software, desktop, and embodied benchmarks stress
realistic interfaces
\citep{zhou2023webarena,koh2024visualwebarena,deng2023mind2web,yao2022webshop,shridhar2020alfworld,yang2023sweagent,xie2024osworld}.
CLIFT differs from prompt-level search because its selector reuses a
certified per-URL bank learned during RL training, not just more samples.

\paragraph{RL credit assignment and process rewards.}
GRPO and related RLVR methods improve LLMs from scalar outcomes
\citep{schulman2017ppo,shao2024grpo,deepseekr1,xu2025webrl}, while
agent variants refine credit assignment through stratification, anchor
states, success rates, or Monte-Carlo values
\citep{zhu2025stratifiedgrpo,feng2025gigpo,chen2025step,kazemnejad2024vineppo}.
Process-verifier and rich-feedback methods provide denser supervision
\citep{cobbe2021training,lightman2023letsverify,wang2024math,hubotter2026sdpo,madaan2023selfrefine,agarwal2024onpolicy},
and LLM judges provide it for open-ended generations
\citep{bai2022rlaif,liu2023geval,zheng2023judging,rafailov2023dpo}.
CLIFT keeps GRPO but changes the reward object: judge feedback is
augmented by a certified self-verifier and normalised by URL state.

\paragraph{Conformal reliability and specification mining.}
Conformal prediction gives finite-sample reliability guarantees
\citep{vovk2005algorithmic,angelopoulos2021gentle,gibbs2021aci,vovk2012mondrian},
with recent adaptations to language modeling, decision making,
off-policy prediction, LLM-agent routing, and LLM judging
\citep{conformalLm2024,lekeufack2024conformaldecision,kuipers2024conformaloffpolicy,ccpo2025,badshah2026scope}.
Those works calibrate outputs, policies, or judgments; CLIFT calibrates
the \emph{verification questions} themselves, using URL scope, polarity,
and a Mondrian-\aci\ trust weight. A CEGAR-style loop
\citep{clarke2000cegar} turns judge rationales into reusable
specifications for both training and test-time selection.

\section{Conclusion}
\label{sec:conclusion}

CLIFT turns expensive judge feedback into a reusable verifier for web
agents: conformally certified URL-state evidence densifies training
reward, then the frozen bank supports judge-free Conformal Trajectory
Selection. Across \wai, \vwa, and \om, the same bank helps train an
open model, transfers to a stronger closed model, and recalibrates for
live-web selection without policy fine-tuning. A trajectory score is
consumed once; a certified bank can be inspected, recalibrated, and
reused.

The bank is the durable artefact behind all three results. Each
certified question carries a URL scope, a polarity sign, and a trust
weight derived from observed lift, with provenance back to the judge
rationale that introduced it. This makes the bank small enough to
ship with the agent, auditable when a result looks off, and
recoverable: when a question's lift collapses on a new distribution,
the certifier withholds trust rather than amplifying noise. Together
these properties give one knob, namely what the judge taught the
agent to notice, that persists from training reward to test-time selection,
from open backbone to closed backbone, and from sandboxed verifiers
to live-web judges, and that stronger future judges can grow without
discarding what earlier rounds taught.

\bibliographystyle{plainnat}
\bibliography{refs}

\appendix
\section{Appendix}
\label{sec:appendix}

This appendix contains evaluation prompts, the CLIFT-only training
setup, detailed bank-transfer accounting for \om{}, and case studies
drawn from evaluation logs.

\subsection{Evaluation Prompts}
\label{sec:evaluation-prompts}

\paragraph{\wai{} browser-use agent prompt.}
The \wai{} browser-use rollout code uses the following system and user
templates (abridged only by replacing runtime fields with braces).
\begin{quote}
\footnotesize\ttfamily\raggedright
You are a browsing agent completing a task on a web app.\\
Each turn you see a simplified accessibility snapshot of the current
page listing interactable elements with numeric ids. Call exactly ONE
tool per turn: click(element\_id), type\_text(element\_id, text,
press\_enter=True), scroll(direction), goto(url), go\_back(),
press(key), hover(element\_id), stop(answer).\\
Do not narrate. Emit a single tool call.\\[2pt]
Task: \{intent\}\\
Current URL: \{url\}\\
Page title: \{title\}\\
Accessibility snapshot: \{snapshot\}\\
Recent action history: \{history\}
\end{quote}

\paragraph{\vwa{} Gemma-4 agent prompt.}
The \vwa{} evaluation harness uses the Gemma-4 browser prompt below.
\begin{quote}
\footnotesize\ttfamily\raggedright
You are an autonomous intelligent agent tasked with navigating a web
browser. You will be given web-based tasks.\\
The user's objective: This is the task you're trying to complete.\\
The current web page screenshot has numbered interactable elements.\\
The observation lists interactable elements as [id] [tagType] [text].\\
Use browser control tools to navigate and interact with pages.\\
Rules: issue only valid actions; issue one tool call at a time; reason
step by step before acting; use element ids; verify before stop; do not
stop prematurely; do not construct internal URLs with goto.\\[2pt]
OBJECTIVE: \{objective\}\\
URL: \{url\}\\
OBSERVATION: \{observation\}\\
PREVIOUS ACTION: \{previous\_action\}
\end{quote}

\paragraph{\om{} agent and WebJudge prompts.}
The \om{} rollout prompt uses the same browser-agent pattern, but for
live websites and final-answer tasks:
\begin{quote}
\footnotesize\ttfamily\raggedright
You are a web browsing agent. You will be given a task objective and
must interact with a real website to complete it.\\
Each turn you receive a screenshot with numbered element IDs, an
accessibility tree, the current URL, and your previous action.\\
Call exactly one tool per turn. Do not construct URLs; click links on
the page. Never stop on the first step. For find/show/look-up tasks,
navigate to the relevant page first. Verify before calling stop.\\[2pt]
OBSERVATION: \{observation\}\\
URL: \{url\}\\
OBJECTIVE: \{objective\}\\
PREVIOUS ACTION: \{previous\_action\}
\end{quote}
The canonical \om{} judge (WebJudge;~\citealp{xue2025illusion}) first extracts key points, scores snapshots,
and then applies the final judging prompt:
\begin{quote}
\footnotesize\ttfamily\raggedright
You are an expert in evaluating the performance of a web navigation
agent. Given the user's task, the agent's action history, key points
for task completion, and important web pages in the trajectory,
determine whether the agent completed the task and achieved all
requirements.\\
Important criteria: filters must be applied correctly; exact ranges
must match; submission/display actions must be completed; invalid or
empty retrieved information may still be success if the required action
was correctly performed.\\
Respond in two lines:\\
Thoughts: \{reasoning\}\\
Status: "success" or "failure"
\end{quote}

\subsection{CLIFT Training Setup}
\label{sec:clift-training-setup}

This subsection documents the training setup for the reported
\method{} checkpoint, organised into the four moving parts of the
system: the backbone and the task pool that drives RL rollouts; the
compute budget and shard layout; the per-rollout reward and
verification loop; and the certification rule that turns the running
bank into the deployment object loaded by CTS at test time. The
concrete hyperparameter values used by all four stages are listed
separately in \S\ref{sec:reported-hparams}.

\paragraph{Backbone and shard layout.}
The reported \method{} checkpoint trains Gemma-4-31B-it with a rank-8
LoRA adapter. We use $16$ single-node shards, each with its own
browser workers, reward pipeline, and LoRA adapter. After training,
the per-shard adapters are averaged element-wise and merged into the
base model for evaluation. \om{} is excluded from policy training
entirely; we use it only for bank transfer and CTS at evaluation.

\paragraph{Seed pool.}
The training pool is synthesised from scratch without reading any
evaluation task. \wai{} seeds ($186$ tasks) are drawn from each
application's schema and user-facing documentation; \vwa{} seeds
($206$ tasks) come from a separate SynthAgent crawl that probes the
three Docker environments via header-only navigation. Both seed sets
describe what the apps \emph{can do}, not what the public benchmarks
\emph{ask the agent to do}.

\paragraph{Frontier-LLM expansion.}
Two frontier LLMs (Claude Opus~4.6 and GPT-5.5) expand each seed into
a draft pool of roughly $20{\times}$ candidates per seed. Each
expansion call sees only the seed's schema or crawl trace and a small
list of formatting constraints; it never sees the public benchmark
task list, evaluator code, or any reference answer.

\paragraph{Four-stage filter.}
A shared filter then removes (i)~\emph{malformed} tasks that fail to
parse or execute against a sandbox replay, (ii)~\emph{near duplicates}
under embedding-based dedup at cosine $0.92$, (iii)~\emph{hardcoded}
tasks whose answers are already present verbatim in the seed schema,
and (iv)~\emph{trivial-navigation} tasks that resolve in a single URL
change without any UI interaction or form input. What remains is the
$1{,}886$ \wai{} and $1{,}272$ \vwa{} training tasks used for all
reported runs.

\paragraph{Train/eval isolation.}
At no point during synthesis, filtering, sharding, or RL training did
the pipeline read \wai{}, \vwa{}, or \om{} evaluation task
descriptions, evaluation HTML/DOM snapshots, or evaluation reward
signals. The eval suites are first loaded only after all training
runs and adapter averaging are complete, and only by the read-only
evaluator.

\paragraph{Compute resources.}
Training ran as shard-parallel single-node jobs on H200 machines; at
peak we used eight nodes, each with $8{\times}$H200 GPUs, and completed
$16$ shard jobs across the \wai{} and \vwa{} training pools. Within each
node, resources were split between vLLM serving, LoRA optimization, and
browser-rollout processes. Closed-backbone \vwa{} and \om{} evaluations
used GPT-5.5 API inference for rollouts and CTS candidates, but did not
fine-tune the closed model.

\paragraph{Rollout and verification loop.}
Each prompt produces a GRPO group of $G=8$ browser rollouts. One
comparative batch judge call scores the group at step level and supplies
reasoning for CEGAR. In parallel, the base Gemma-4 model answers
verification questions with the LoRA adapter detached, so verifier
answers come from the same backbone without using the trained task
adapter.

\paragraph{Certification, reward, and deployment object.}
For each question, CCC maintains global, host, and URL-path
Mondrian-ACI trackers with target $\alpha^\star=0.10$ and lift-based
trust weights; certification uses $n_{\min}=50$ and
$\mathrm{lift}_{\min}=0.02$. Certified verifier evidence is added
through the asymmetric per-step blend, so it can support the judge
reward but cannot subtract from it. The resulting URL-stratified returns
are written to token advantages over tool-call spans, and the frozen
certified bank is the deployment object reused by CTS at test time.

\subsection{Reported CLIFT Hyperparameters}
\label{sec:reported-hparams}

The method section uses symbolic quantities; the concrete values used
for the reported runs are: judge threshold $\theta_j=0.5$; GRPO group
size $G=8$; certification target $\alpha^\star=0.10$; minimum
certification count $n_{\min}=50$; minimum lift
$\mathrm{lift}_{\min}=0.02$; CEGAR interval $M=60$ training steps;
CEGAR batch size 3 to 5 questions; reward blend $\beta=0.15$; return
discount $\gamma=0.95$; CTS vote count $m=3$ with self-verifier
temperature $0.7$; greedy incumbent temperature $0$; diverse retry
temperature $1.0$; and seed-bank sizes $281$ \wai{} questions and $89$
\vwa{} questions before filtering and certification.

\subsection{Bank Details and OM2W Transfer}
\label{sec:bank-transfer-appendix}

This appendix tracks the construction, certification, and transfer of
the verification banks that drive both training and test-time
selection. For each benchmark we report the seed bank size, the size
of the candidate bank after CEGAR-driven growth across training
shards, the certified subset retained by the conformal certifier,
and (for \om{} only) the subset that survives filtering and rewriting
for live-web reuse. The same bank object is read by the training
reward (\S\ref{sec:training}) and by CTS at test time
(\S\ref{sec:cts}); the numbers below describe how that object grows
and shrinks at each stage.

\paragraph{\wai{} and \vwa{} training banks.}
The \wai{} training bank starts with $281$ seed verification
questions derived from app schemas and user documentation; the \vwa{}
training bank starts with $89$ seed questions from SynthAgent site
exploration of the three Docker environments. Both banks grow
through the CEGAR refinement loop, in which the judge proposes $3$ to
$5$ new questions every $M{=}60$ training steps targeting rollouts
that the existing bank fails to discriminate, and shrink through
certification: questions whose lift falls below $0.02$ at all tiers,
or whose polarity is contradicted by observed lift, are demoted to
neutral or dropped. After cross-shard merging, the \wai{} candidate
bank totals $992$ unique questions, of which $44$ retain non-zero
trust weights at one or more URL tiers and form the certified \wai{}
bank used by the training reward and by CTS at test time. The
\vwa{} candidate bank totals $1{,}565$ unique questions with $21$
certified at the end of training. The $36$ \wai{} and $48$ \vwa{}
questions reported in Table~\ref{tab:appendix-om2w-bank} are the
portion of each bank that the filter/rewrite stage judged general
enough to reuse on live-web \om{}: certified questions kept as-is
($7$ \wai{}, $10$ \vwa{}) plus rewritten variants of certified or
promising uncertified questions ($29$ \wai{}, $38$ \vwa{}). The
remainder stay in their original training banks because their URL
patterns or DOM references are app- or environment-specific and do
not transfer.

\paragraph{Representative certified questions.}
To make the bank concrete, we walk through six representative
certified questions drawn from the merged \wai{} and \vwa{} banks,
using the same prompt-style format as the agent and judge templates
above. Each block shows the question identifier, the URL pattern at
which it applies (\texttt{*} = global; the bank also contains
per-host and per-URL-path refinements such as \texttt{*google.com*}
and \texttt{*reddit.com/r/*} that share the same text but track
separately under Mondrian-\aci{}), the observed trust weight $w$
(signed by polarity, magnitude scaled by lift), and the number of
judge-scored observations $n$ that supported certification, followed
by the YES/NO question that is asked of every rollout at runtime.

\Needspace{12\baselineskip}
\paragraph{Q.1: Reaching content-bearing pages
(\wai, $w{=}{+}0.54$).}
The single strongest signal in the bank. Rollouts that visit a page
containing task-relevant content are far more likely to receive a
high judge score than rollouts that loiter on app homepages or
navigation menus, and this pattern holds across all 9 \wai{} apps.
\begin{quote}
\footnotesize\ttfamily\raggedright
Identifier: made\_meaningful\_progress\_toward\_goal\\
URL pattern: *\\
Trust weight: $w = +0.54$, observations: $n = 82$.\\[2pt]
Did the agent reach at least one page or state that contains
information directly relevant to answering or completing the task
(not just the app homepage or navigation menus)?
\end{quote}

\Needspace{12\baselineskip}
\paragraph{Q.2: Constructive non-destructive behaviour
(\wai, $w{=}{+}0.54$).}
A two-part check that distinguishes "did nothing harmful" from "did
something productive". Rollouts that take destructive shortcuts
(delete, cancel, revoke) without a constructive counterpart fail this
question even when the final URL looks plausible.
\begin{quote}
\footnotesize\ttfamily\raggedright
Identifier: avoided\_destructive\_unintended\_actions\\
URL pattern: *\\
Trust weight: $w = +0.54$, observations: $n = 82$.\\[2pt]
Did the agent both (a) avoid performing destructive actions (delete,
cancel, revoke) not required by the task AND (b) take at least one
constructive action toward the task goal?
\end{quote}

\Needspace{12\baselineskip}
\paragraph{Q.3: Source-grounded final answer
(\wai, $w{=}{+}0.27$).}
Targets a common failure mode in information-retrieval tasks:
emitting an answer without ever opening the page that actually
contains the requested fact. Certified at high $n$, so the verifier
has stable evidence at the global tier.
\begin{quote}
\footnotesize\ttfamily\raggedright
Identifier: extracted\_info\_from\_correct\_page\\
URL pattern: *\\
Trust weight: $w = +0.27$, observations: $n = 520$.\\[2pt]
For information-retrieval tasks, did the agent extract its final
answer from the page that actually contains the requested information?
\end{quote}

\Needspace{12\baselineskip}
\paragraph{Q.4: End-state coherence
(\wai, $w{=}{+}0.26$).}
A simple URL coherence check that catches "wandered off" rollouts in
which the agent ends on an unrelated section of the app. Cheap to
answer for the base verifier (it only needs the final URL and the
task description) and stable across apps.
\begin{quote}
\footnotesize\ttfamily\raggedright
Identifier: final\_page\_matches\_task\_context\\
URL pattern: *\\
Trust weight: $w = +0.26$, observations: $n = 238$.\\[2pt]
Is the final URL or page the agent ended on relevant to the task
objective rather than an unrelated section of the app?
\end{quote}

\Needspace{12\baselineskip}
\paragraph{Q.5: Active search behaviour
(\vwa, $w{=}{+}0.20$).}
A \vwa-specific predicate. Distinguishes agents that submit a query
and then act on the results from agents that issue a search and
freeze on the results page without clicking through. Especially
discriminative on the Classifieds and Shopping environments.
\begin{quote}
\footnotesize\ttfamily\raggedright
Identifier: used\_search\_and\_acted\_on\_results\\
URL pattern: *\\
Trust weight: $w = +0.20$, observations: $n = 66$.\\[2pt]
Did the agent both submit a search query AND subsequently take a
meaningful action on the results (e.g., clicking a result link rather
than just observing the search page)?
\end{quote}

\Needspace{12\baselineskip}
\paragraph{Q.6: Navigation depth
(\vwa, $w{=}{+}0.13$).}
Many \vwa{} tasks require going past the search-results page to a
specific listing or product detail page to extract a price, image, or
description. This question rewards depth of navigation rather than
breadth and is the highest-trust \vwa{} question whose URL pattern is
already general enough to translate to live-web \om{} without rewriting.
\begin{quote}
\footnotesize\ttfamily\raggedright
Identifier: visited\_target\_detail\_page\\
URL pattern: *\\
Trust weight: $w = +0.13$, observations: $n = 200$.\\[2pt]
Did the agent navigate to a specific item detail page (product page,
listing page, or submission page) relevant to the task, beyond the
search results page?
\end{quote}

These six illustrate three properties the certifier rewards: the
question must discriminate between high- and low-judge rollouts
(positive lift), it must be answerable from the trajectory by the
base self-verifier (not too low-level for the LoRA-detached Gemma-4
to read off), and its YES/NO must align with the judge's separation
under the chosen polarity. Questions that fail these tests, e.g.,
ones that ask about UI elements specific to a single page or that
the base model cannot reliably answer, are demoted to neutral or
dropped by the polarity-veto rule of Eq.~\ref{eq:polarity-veto}.

\paragraph{\om{} transfer setting.}
The main text reports the compact bank-transfer table. Here we keep
the more detailed accounting for how the \om{} bank was produced and
why it is a strict transfer setting. No reported policy is trained on
\om{}; only the question bank is filtered, rewritten, expanded, and
recalibrated.

\begin{table}[t]
\centering
\caption{\textbf{\om{} bank construction and recalibration.} Questions
come from \wai/\vwa{} banks plus an \om{}-specific failure-mode
expansion.}
\label{tab:appendix-om2w-bank}
\footnotesize
\setlength{\tabcolsep}{6pt}
\renewcommand{\arraystretch}{1.02}
\begin{tabular}{@{}llrrr@{}}
\toprule
Origin & Stage & Questions & Certified & Mean lift \\
\midrule
\wai{} bank & kept as-is & 7 & 7 & 0.46 \\
\wai{} bank & rewritten & 29 & 26 & 0.46 \\
\vwa{} bank & kept as-is & 10 & 4 & 0.31 \\
\vwa{} bank & rewritten & 38 & 27 & 0.28 \\
\om{} expansion & live-web checks & 48 & 36 & 0.30 \\
\midrule
\textbf{Total} & -- & \textbf{132} & \textbf{100} & \textbf{0.34} \\
\bottomrule
\end{tabular}
\end{table}

\paragraph{Transferred question taxonomy.}
The filter/rewrite stage keeps general web-interaction predicates
where possible and removes benchmark-specific app names or URL
patterns when they do not transfer. The expansion stage adds questions
for live-web failure modes observed on \om{}, including bot-block
pages, timeout or wandering behaviour, hallucinated extraction,
partial-filter application, search-vs-detail-page confusion, and
wrong-final-answer stops. The resulting 132-question bank uses the
same schema as the training-time bank, so CTS loads it without code
changes.

\paragraph{Recalibration.}
Each question is scored on a stratified pass/fail rollout sample using
the base Gemma-4 verifier with logprob-extracted YES probabilities.
Lift is computed as the pass/fail separation of the YES probability.
Questions whose absolute lift is below the discrimination floor remain
in the bank with zero weight; they can be re-certified later on a
different rollout sample. Among the 100 certified \om{} questions, 75
have positive lift and 25 have negative lift, so the selector receives
both progress evidence and failure-mode evidence.

\subsection{OM2W Transfer Performance Details}
\label{sec:om2w-transfer-performance-appendix}

\begin{table}[t]
\centering
\caption{\textbf{\om{} transfer performance details.} $K$ is the rollout
budget and every row emits one selected trajectory. Canonical judge is
WebJudge~\citep{xue2025illusion} + o4-mini at threshold 4; Sonnet is a cross-check.}
\label{tab:appendix-om2w-performance}
\footnotesize
\setlength{\tabcolsep}{8pt}
\renewcommand{\arraystretch}{1.02}
\begin{tabular}{@{}lrrrl@{}}
\toprule
System & $K$ & Canonical SR & Gain & Sonnet SR \\
\midrule
Gemma-4 base & 1 & 40.0 & -- & 46.8 \\
Gemma-4 + ours & 2 & 45.0 & +5.0 & partial positive \\
GPT-5.5 low base & 1 & 43.7 & -- & 47.3 \\
GPT-5.5 high base & 1 & 49.7 & -- & 52.0 \\
GPT-5.5 high + ours & 2 & 57.3 & +7.7 & 55.7 \\
GPT-5.5 high + ours & 4 & 61.0 & +11.3 & 61.7 \\
\bottomrule
\end{tabular}
\end{table}

Table~\ref{tab:appendix-om2w-performance} reports the detailed
cross-judge performance numbers behind the compact main-text \om{}
figure. It is intentionally separate from
Table~\ref{tab:appendix-om2w-bank}: the bank table explains how the
question set was transferred and recalibrated, while this table measures
how much realised success that transferred evidence adds after CTS
selects one trajectory to emit.

The canonical column is the number used in the main paper, since it
matches the \om{} judging protocol. The Sonnet column is included only
as an external consistency check. It follows the same ordering at the
important operating points: the transferred selector improves the
Gemma-4 backbone, and the GPT-5.5 high-effort setting gains further as
the rollout budget increases from $K{=}2$ to $K{=}4$. Rows with $K{=}1$
are ordinary single rollouts; rows with larger $K$ are CTS-selected
outputs, not oracle unions over multiple attempts.

This table closes the quantitative \om{} appendix. The next subsection
switches from aggregate accounting to qualitative trajectory evidence.

\subsection{Case Studies}
\label{sec:case-studies}

\paragraph{Case A.1: CTS save on \wai{} Gmail.}
The selector replaces a long failing rollout with a shorter successful
one on the same task.
\begin{quote}
\footnotesize\ttfamily\raggedright
Task: Change profile name to "Alex J. Johnson"; create label "Priority"
with color \#EA4335; apply it to all contacts in the Family label.\\
Greedy trajectory: 87 browser steps; final verifier label = fail.\\
Diverse trajectory: 25 browser steps; changed profile name, created the
Priority label, applied it to Laura Johnson-Martinez, Leo Martinez,
Margaret Johnson, and Richard Johnson; final verifier label = pass.\\
CTS decision: swap=True, greedy score=-0.5, diverse score=+0.5,
chosen\_success=True.
\end{quote}

\paragraph{Case A.2: CTS save on a long \wai{} Handshake chain.}
The selected rollout completes a multi-hop message, event, and job chain
that the greedy rollout leaves unfinished.
\begin{quote}
\footnotesize\ttfamily\raggedright
Task: Read the latest unread message, RSVP to that sender's upcoming
event, and save all active jobs from that sender.\\
Greedy trajectory: failed to complete the full message/event/job chain.\\
Diverse trajectory: navigated through messages, identified the sender's
event, RSVP'd, and saved the relevant active jobs.\\
CTS decision: swap=True, greedy score=-0.5, diverse score=+0.5,
chosen\_success=True.
\end{quote}

\paragraph{Case A.3: No-regression behaviour on \wai{} Gmail.}
CTS can also decline a replacement when the greedy rollout is already
the better candidate.
\begin{quote}
\footnotesize\ttfamily\raggedright
Task: Disable two-factor authentication.\\
Greedy trajectory: completed the account-security change successfully.\\
Diverse trajectory: failed the task.\\
CTS decision: swap=False; the selector kept the greedy rollout rather
than replacing a working trajectory with a worse sample.
\end{quote}

\paragraph{Case A.4: Live-web recovery on \om{} Indeed.}
On a live site, the successful trajectory recovers from blank results
and an additional-verification page by returning to the native search
flow.
\begin{quote}
\footnotesize\ttfamily\raggedright
Task: Find support-services jobs in Bentonville, Arkansas.\\
Trajectory snippet: navigate to Indeed; enter "support services" and
"Bentonville, AR"; direct results URL first renders blank; alternate URL
hits an additional-verification page; agent returns home, uses the native
search form, waits for results, and stops with visible listings.\\
Final answer: includes Healthcare Driver in Bentonville and other
support-services listings; WebJudge label = success.
\end{quote}

\end{document}